\documentclass[10pt,twocolumn,letterpaper]{article}
\usepackage{ijcb}
\usepackage{times}
\usepackage{epsfig}
\usepackage{graphicx}
\usepackage{amsmath}
\usepackage{amssymb}
\usepackage{subcaption}
\usepackage{tabularx}
\usepackage{threeparttable}
\usepackage{multirow}
\usepackage{comment}
\usepackage{makecell}
\usepackage{enumitem}
\usepackage{enumerate}
\usepackage{MnSymbol}
\newcolumntype{L}{>{\raggedright\arraybackslash}X}

\usepackage[pagebackref=true,breaklinks=true,letterpaper=true,colorlinks,bookmarks=false]{hyperref}

\ijcbfinalcopy 

\ifijcbfinal
\begin{document}

\title{Liveness Detection Competition - Noncontact-based Fingerprint\\ Algorithms and Systems (LivDet-2023 Noncontact Fingerprint)}

\author{Sandip Purnapatra$^1{}^*{}^\dag$, Humaira Rezaie$^1{}^\dag$, Bhavin Jawade$^2{}^\dag$, Yu Liu$^1{}^\dag$, Yue Pan$^1{}^\dag$, Luke Brosell$^1{}^\dag$,\\ Mst Rumana Sumi$^1{}^\dag$, Lambert Igene$^1{}^\dag$, Alden Dimarco$^1{}^\dag$, 
Srirangaraj Setlur$^2{}^\dag$, 
Soumyabrata Dey$^1{}^\dag$,\\
Stephanie Schuckers$^1{}^\dag$,\\
Marco Huber$^3{}^\ddag$, Jan Niklas Kolf$^3{}^\ddag$, Meiling Fang$^3{}^\ddag$, Naser Damer$^3{}^\ddag$, Banafsheh Adami$^4{}^\ddag$, Raul Chitic$^5{}^\ddag$,\\ Karsten Seelert$^5{}^\ddag$, Vishesh Mistry$^6{}^\ddag$, Rahul Parthe$^6{}^\ddag$, Umit Kacar$^6{}^\ddag$ 
\and
\small$^1${Clarkson University}, $^2${University at Buffalo}, $^3${Fraunhofer IGD}, $^4${West Virginia University}, $^5${Dermalog}, $^6${Tech5}, $\dag${Organizers}, $\ddag${Competitors}
\and 
\small$^*$Corresponding email: purnaps@clarkson.edu
}
\maketitle
\begin{abstract} 
Liveness Detection (LivDet) is an international competition series open to academia and industry with the objective to assess and report state-of-the-art in Presentation Attack Detection (PAD). LivDet-2023 Noncontact Fingerprint is the first edition of the noncontact fingerprint-based PAD competition for algorithms and systems. The competition serves as an important benchmark in noncontact-based fingerprint PAD, offering (a) independent assessment of the state-of-the-art in noncontact-based fingerprint PAD for algorithms and systems, and (b) common evaluation protocol, which includes finger photos of a variety of Presentation Attack Instruments (PAIs) and live fingers to the biometric research community (c) provides standard algorithm and system evaluation protocols, along with the comparative analysis of state-of-the-art algorithms from academia and industry with both old and new android smartphones. The winning algorithm achieved an APCER of 11.35\% averaged over all PAIs and a BPCER of 0.62\%. The winning system achieved an APCER of 13.0.4\%, averaged over all PAIs tested over all the smartphones, and a BPCER of 1.68\% over all smartphones tested. Four-finger systems that make individual finger-based PAD decisions were also tested. The dataset used for competition will be available \footnote{\url{https://noncontactfingerprint2023.livdet.org/index.php}} to all researchers as per data share protocol. 

\end{abstract}

\section{Introduction}
Biometrics is replacing the traditional, inconvenient and insecure password/PIN authentication worldwide and offering a modern, convenient, and secure biometric recognition solution. Among the biometric modalities, the fingerprint is universal and offers a high level of accuracy, uniqueness, and permanence. Thus, making it widely used and a popular biometric modality with applications in industries, law enforcement agencies, and national ID programs worldwide \cite{lin2018matching}. ``Recent research has revealed that contact-based nature and the quality differences of these sensors can lead to various types of performance issues \cite{labati2015toward}, such as:
\begin{itemize}[noitemsep,leftmargin=*]
\item {\bf Sensors hard to clean and sterilize} - Latent fingerprints of a subject can interfere with the following captures
\item {\bf Distortion of fingers from touching sensor surface} - Elastic deformation  can be caused by the friction between finger skin and sensor surface during fingerprint collection and thus result in a limited performance
\item {\bf Capture problems from certain challenges} - Skin deformation due to age or injury, humidity can cause low capture contrast and the failure-to-capture
\item {\bf No universal sensor} - Wide range of fingerprint sensors are available worldwide and most are used as an accessory to a device. The inclusion of a secure fingerprint sensor in a smartphone raises the cost of the product.
\end{itemize}
Noncontact fingerprint-based systems eliminate the necessity of additional biometric sensors, lowering the cost of smartphones, and most importantly still keeping the integrity of smartphone security against a breach of confidentiality or sensitive data leakage \cite{fujio2018face} \cite{marasco2021fingerphoto}. However, noncontact-based systems must be frequently tested against ever-evolving presentation attacks. \par
LivDet-2023 Noncontact Fingerprint Algorithm and System is an international competition and the first noncontact fingerprint-based liveness detection competition of the LivDet series to test the state-of-the-art noncontact-based fingerprint PAD with an independent evaluation of the submitted algorithms and systems with known and unseen presentation attacks. LivDet-2023 Noncontact Fingerprint System competition was conducted for four-finger-based noncontact fingerprint systems and did not offer any official training dataset-- the competitors were free to use any proprietary and/or publicly available data to train their Systems.  LivDet-2023 Noncontact fingerprint Algorithm competition was conducted for single fingertip-based algorithms and did offer training data to the participants to focus on the standardization of the evaluation capabilities of the state-of-the-art algorithms. \par
The most significant contributions of this publication and the LivDet-2023 Noncontact Fingerprint Algorithm and System competition are: 

\begin{itemize}[leftmargin=*,noitemsep]
\item A report on the state-of-the-art in noncontact-based fingerprint PAD on independent testing of {\bf four algorithms submitted} and {\bf three noncontact fingerprint applications submitted as systems} to the competition organizers. 

\item Dataset prepared in accordance to Fast Identity Online (FIDO) Biometric Requirements \cite{schuckers2023fido} and all submitted algorithms and systems were evaluated by standard PAD metrics as defined by International Organization for Standardization
(ISO) \cite{ISO_IEC_301073:2017}.

\item  Largest spectrum of PAIs used to date, to the best of our knowledge, and used for the noncontact fingerprint competition. {\bf Six different PAIs} constitute the test dataset with each PAI category captured with multiple different smartphone (old and new) back cameras as sensors.

\item Testing noncontact fingerprint detection algorithms against of {\bf two novel PAIs}: latex PAI and, synthetically generated fingertips PAI, created from live subjects to replicate real-life uncertain scenarios

\item {\bf Initiation of LivDet-2023 Noncontact Fingerprint Algorithm and System competition}. The competition benchmarks the testing protocols and will be available to all researchers after the competition is concluded, to allow testing of all future algorithms and systems with the LivDet-2023 Noncontact fingerprint algorithm and system competition results. 
\end{itemize}

\section{Performance Evaluation Metrics}
\label{evaluation}

LivDet-2023 Noncontact Fingerprint Algorithm and System competition employs two basic PAD metrics for evaluation which follow the recommendations of ISO/IEC 30107-3 \cite{ISO_IEC_301073:2017}:
\begin{itemize}[leftmargin=*, noitemsep]
\item\textbf{Attack Presentation Classification Error Rate (APCER)}, the proportion of attack presentations of the same PAI species incorrectly classified as bonafide presentation, or PAI classified as live. 
\item\textbf{Bonafide Presentation Classification Error Rate (BPCER)}, the proportion of bonafide presentations classified as attack presentations, or live classified as PAI.\end{itemize}

Both the APCER and BPCER metrics are used to evaluate the algorithms. ISO also recommends using the maximum value of APCER when multiple PA species (or categories) are present in the case of system-level evaluation, which is primarily designed for industrial applications. For this competition, our goal is to consider the detection of all PAIs, and not to rank the algorithms submitted by the competitors from the worst - to the best-performing PA. Thus, in the LivDet-2023 Noncotact Fingerprint competition, we evaluated performance based on the weighted average of APCER over all PAIs:

\begin{itemize}[leftmargin=*, noitemsep]
\item \textbf{Weighted Average of APCER} (APCER$_{\mbox{\footnotesize average}}$), which is the average of APCER over all PAIs and weighted by the number of samples in each PAI category.

\noindent
For the {\bf purpose of competition ranking}, the Average Classification Error Rate (ACER) was computed to select the best performer

\item \textbf{Average Classification Error Rate (ACER):} the average of APCER$_{\mbox{\footnotesize average}}$ and BPCER.
\end{itemize}
Note that ACER has been deprecated in ISO/IEC 30107-3:2017 \cite{ISO_IEC_301073:2017} in the industry-related PAD evaluations.

\section{Noncontact Fingerprint PAD efforts so far}
Many researchers have been exploring smartphone-based finger photo recognition in the last decades \cite{lee2006preprocessing} \cite{lee2008recognizable} \cite{li2012testing} \cite{sankaran2015smartphone} \cite{jawade2021multi} \cite{jawade2022ridgebase}. Most of these works were focused on finger photo-processing techniques i.e. segmentation, enhancement for finger photo quality improvements, and conversion from finger photos to digital fingerprints. These converted digital fingerprints were then used minutia-based matching algorithms to demonstrate improvement in authentication performances. These researches demonstrated the challenges of finger photo processing, however, the corresponding dataset were not made available publicly to encourage further research by many of the studies. Early in the last decade, Stein et al. \cite{stein2013video} developed a PAD algorithm for PAI detection in a smartphone-based noncontact fingerprint-capturing application. Another work from the authors of Taneja et al. \cite{taneja2016fingerphoto} studied finger photo PAD on mobile devices with print-out and replay attacks. The PAD dataset is publicly available, however, the manual capture of live and finger photos was not properly focused and thus of low quality and noisy, and neither the data was collected using standard international collection protocols. Fujio et al. \cite{fujio2018face} used the same dataset and developed a PAD algorithm based on CNN (AlexNet) architecture. The same PAD dataset, along with collected replay attack data, was used by the authors Marasco et al. \cite{marasco2021fingerphoto} to develop ResNet-based PAD algorithm. Wasnik et al. \cite{wasnik2018presentation} published PAD-based finger photo recognition. \par
Bhavin et. al. \cite{jawade2021multi} \cite{jawade2022ridgebase} developed Ridgebase which is a large-scale dataset of contactless and contact-based finger photos acquired through user-operated smartphones. The dataset consists of more than 15,000 contactless and contact-based fingerprint image pairs acquired from 88 individuals under different background and lighting conditions using two smartphone cameras and one flatbed contact sensor. However, the dataset was not developed for PAD. Recently, Kolberg et al. \cite{kolberg2023colfispoof} introduced the {\bf COLFISPOOF} dataset for non-contact fingerprint PAD. The dataset has 7200 samples of 72 different PAI species and was captured using two different smartphones. The PAI materials used by the authors are mostly silicon-based and of FIDO level A \& B difficulty, which translates to easy and moderately difficult to make. Furthermore, the authors used {\bf SynCoLFinGer} \cite{priesnitz2022syncolfinger}, which synthetically simulates and generates finger photos from contact-based fingerprints. These synthetic live fingerprints can be very easily visually distinguished from the live fingertips. The authors also proposed evaluation protocols to train and test the PAD algorithms using the developed dataset but did not contribute any PAD algorithm. 

From the evaluation of the state-of-the-art, evidently, the studies that have shared data sets publicly do not have a large spectrum of PAIs, made in accordance with FIDO. The datasets that can be found in the public domain are of low quality, not in focus. None of the studies used extremely sophisticated, hard-to-make, and visually indistinguishable synthetic PAIs to test their developed algorithms. Neither the algorithms were tested against unseen or unknown PAI varieties to replicate real-life uncertain scenarios. Our previous work \cite{purnapatra2023presentation} developed a PAD dataset according to the standard PAI creation protocols, and of three different difficulty levels, using different types of materials and PAI textures that reflect real skin tones. There was scope for further improvement in the performance of our models, we realized the absence of any standard comparison of models and their results across academia and industry. Thus, we hosted the LivDet-2023 Noncontact Fingerprint Algorithm and System competition and invited participants across the industry and academia to participate and compare performances. We used the \cite{purnapatra2023presentation} publication dataset and further enhanced the number of samples with additional PAI collection. System competitors provided an end-to-end system that included capture, user interface, and PAD system, thus we tested with a common set of PAIs and live subjects for all competitors. We recollected the PAI samples for the in-person systems testing for the competition and tested all submitted systems against good quality PAIs of different FIDO PAI levels and live samples from live subjects. 

\subsection{LivDet-2023 Noncontact Fingerprint Algorithm and System competition}
The LivDet-2023 Noncontact Fingerprint Algorithm and System competition is the first LivDet competition on noncontact fingerprint-based PAD and is co-organized by two institutes, namely: Clarkson University (USA), and the University at Buffalo (USA). Previously, LivDet has organized many liveness detection competitions for fingerprint, face, and iris, more details can be found in \cite{LivDet}. The objective of the competition was to evaluate the performance of the state-of-the-art noncontact fingerprint-based PA detection algorithms and systems against traditional and novel PAIs. The competition had two categories: \textit{Algorithms}, and \textit{Systems}. Competitors were given the chance to participate in both the algorithms and systems categories of the competition. International academic and industrial institutions were encouraged to participate in the competition. For the LivDet-2023 Noncontact Fingerprint System competition no official training dataset was offered -- the competitors were free to use any proprietary and/or publicly available data to train their Systems.  The LivDet-2023 Noncontact Fingerprint Algorithm competition offered training data to the participants to focus on the standardization of the evaluation capabilities of the state-of-the-art algorithms {\bf from generalized to uncertain circumstances}. \par
The Algorithm competition was announced for single-fingertip-based algorithms. For the algorithms category of the competition, there were six PAI types - finger photo printout in glossy paper, ecoflex, playdoh, wood glue, latex, and high-quality synthetically generated fingertip photos. A training dataset was shared with the participants containing live and PAIs except for latex and synthetic fingertip. These two PAI categories were kept as unknown modalities. Four submissions from industry and academia were received by the deadline. \par
The System competition was hosted for four-finger-based noncontact fingerprint systems. For the systems category of the competition, there were five PAI types i.e. finger photo printout in glossy paper, ecoflex, playdoh, and wood glue. However, the organizers received five submissions from two participants in this category. Participants sent their noncontact fingerprint-based application software (no hardware with software pre-installed in a smartphone) developed for Android operating systems for evaluation. There were two different types of noncontact Fingerprint-based systems evaluated: applications that make the PAD decision based on each fingertip (individual apps) in a four-finger setting (excluding thumbs of both hands) and the application which makes the PAD decision based on all four fingers of the hand (unified apps). We received three unified apps from two competitors in the systems category. We received two individual apps from a competitor in this category, their evaluations were reported separately in Table: \ref{table:Noncontact_Systems_PAD competition results: Individual Systems category} because there was only one competitor in this category. All the submitted systems were tested against a new Samsung Galaxy A71 smartphone model with Android OS 13 and a few years old Samsung Galaxy S9 model with Android OS 10, to evaluate and compare the performances of the submitted systems which were all Android applications. The goal was to understand the universality of the noncontact fingerprint-based system's performance with one relatively old smartphone back-camera sensor and a relatively better equipped and most recent smartphone back-camera sensor.

\section{Experimental Protocol}

\subsection{ Algorithms and Systems} 
All international academic and industrial organizations were welcome to participate anonymously or non-anonymously in the LivDet-2023 Noncontact Fingerprint Algorithm and System competition. Competitors were given the opportunity to participate in the publication as co-authors. A total of eight teams registered for the competition from across the globe. The competition received four submissions for the algorithms category and two submissions for the systems category.

\subsection{Dataset }
The noncontact fingerprint-based (finger photo) PAD dataset was constructed from 35 live subjects, Additionally, finger-molds  of four fingers of both hands were collected to prepare PAIs. The live data was collected using 6 smartphone pairs. The finger mold (made from a dental mold and impression material) was used to create finger molds or PAIs using five different types of materials, additives to achieve human skin tones with ecoflex PAI, and deep-fake fingertips or synthetic fingers. Table: \ref{Dataset} describes the total number of single-finger live or PAI images present in the LivDet-2023 Noncontact Fingerprint Algorithm competition.  All the four-finger images were segmented to create the single fingertip-based noncontact fingerprint PAD dataset for the competition. Further details about the collection and the dataset pre-processing are available in the previous publication \cite{purnapatra2023presentation}.
\subsection{Presentation Attack Instrument Generation }
The PAIs were prepared using five different types of materials and used to test the submitted noncontact fingerprint systems. All the submitted systems were tested with PAIs created from 20 different identities or the finger molds collected from the 20 different live individuals. Because of the differences in the PAD decision-making of the individual and unified apps, different testing sequences were followed and different test labels to keep track of the testing processes for each type of system were created for testing. The submitted individual apps were tested with a mix of PAI and live fingers (in a four-finger setting, so some fingers with PAI and some without PAI), as the apps make decisions on the individual finger level. For each four-finger setting, the collectors collected four mixed sequence captures with live fingers and PAIs laid on fingertips to test the individual apps. The unified apps were tested with PAIs laid on four fingertips and captured twice for each hand in a four-finger setting with each unified app to get the final test results. Additionally, the individual apps were tested with 14 live subjects in addition to the live tests of the mixed sequence testing. While the unified apps were tested with 28 live subjects with each hand of the subject tested five times per unified app. Testing processes were manual and all the systems' final results were analyzed after quality checks.  More details about the number of tests performed per submitted individual and unified systems are available in Table: \ref{Dataset_Systems}. Additional details about each PAI species are provided as follows:
\begin{itemize}[noitemsep,leftmargin=*]
    \item {\bf Ecoflex layover (EL):} Fingertip layover molds for each finger of the subject's hand mold were prepared using the ecoflex with makeup additives to achieve colors close to human skin tones and used as PAI of difficulty level B. Ecoflex PAI images of 20 subjects were collected.
    \item {\bf Playdoh layover (PL):} Fingertip layover molds for each finger of the subject's hand mold were prepared using playdoh material of red color (easily available in markets) and used as a PAI of difficulty level B. Playdoh PAI images of 20 subjects were collected.
    \item {\bf Wood glue layover (WL):} Fingertip layover molds for each finger of the subject's hand mold were prepared using wood glue. Wood glue PAI images of all 20 subjects were collected and used as PAI of difficulty level B.
    \item {\bf Latex layover (LL):} Fingertip layover molds for each finger of the subject's hand mold were prepared using latex, a rubberized liquid material available for preparing props and prop molds. Latex PAI images of 20 subjects were collected and used as a PAI of difficulty level B to test the performance of the submitted systems.
    \item {\bf Printed PAI (PP):} Four-finger photos were collected from both hands of 20 live subjects and printed out on glossy photo paper using a color-jet printer. This PAI modality has been used as an unseen PAI of level A difficulty to test the performance of the submitted systems.
\end{itemize}
\vspace{-5pt}
\begin{figure*}[!ht]
\centering
\small
\subcaptionbox{\centering \small Printed Finger photo}{\includegraphics[height=3cm,width=0.12\textwidth]{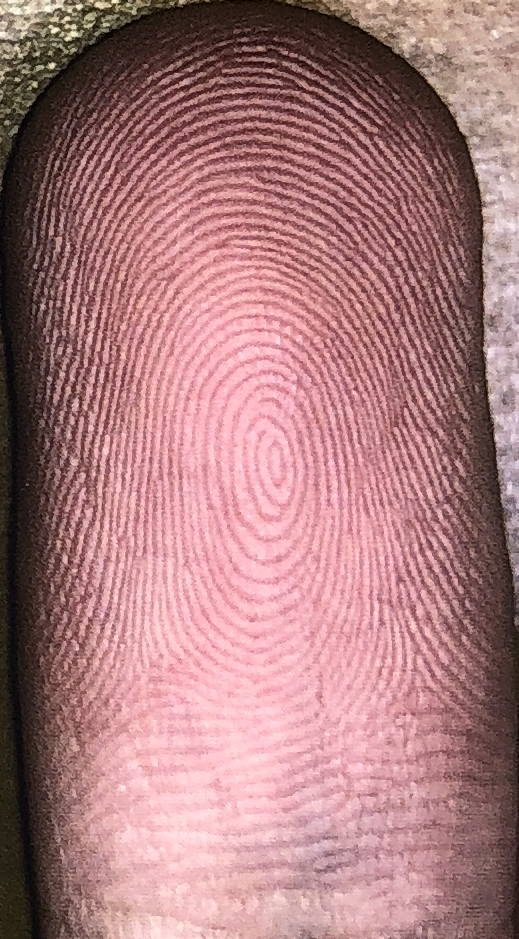}}
\hfill
\subcaptionbox{\centering \small Ecoflex PAI}{\includegraphics[height=3cm,width=0.16\textwidth]{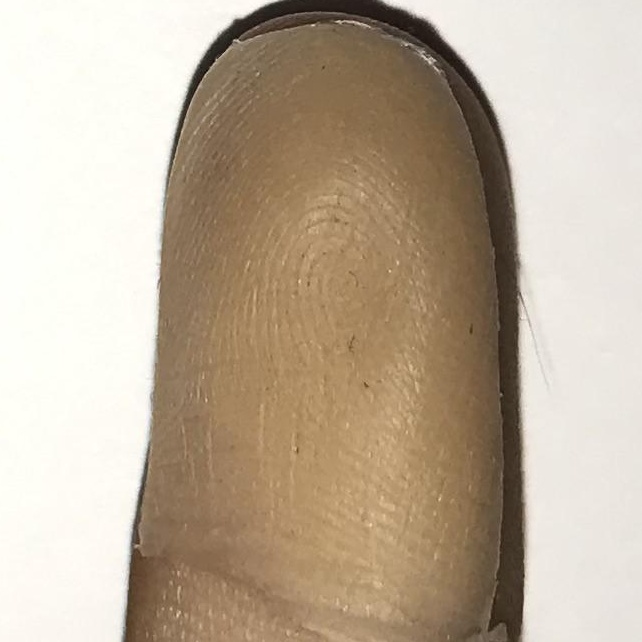}}
\hfill
\subcaptionbox{\centering \small Latex PAI}{\includegraphics[height=3cm,width=0.16\textwidth]{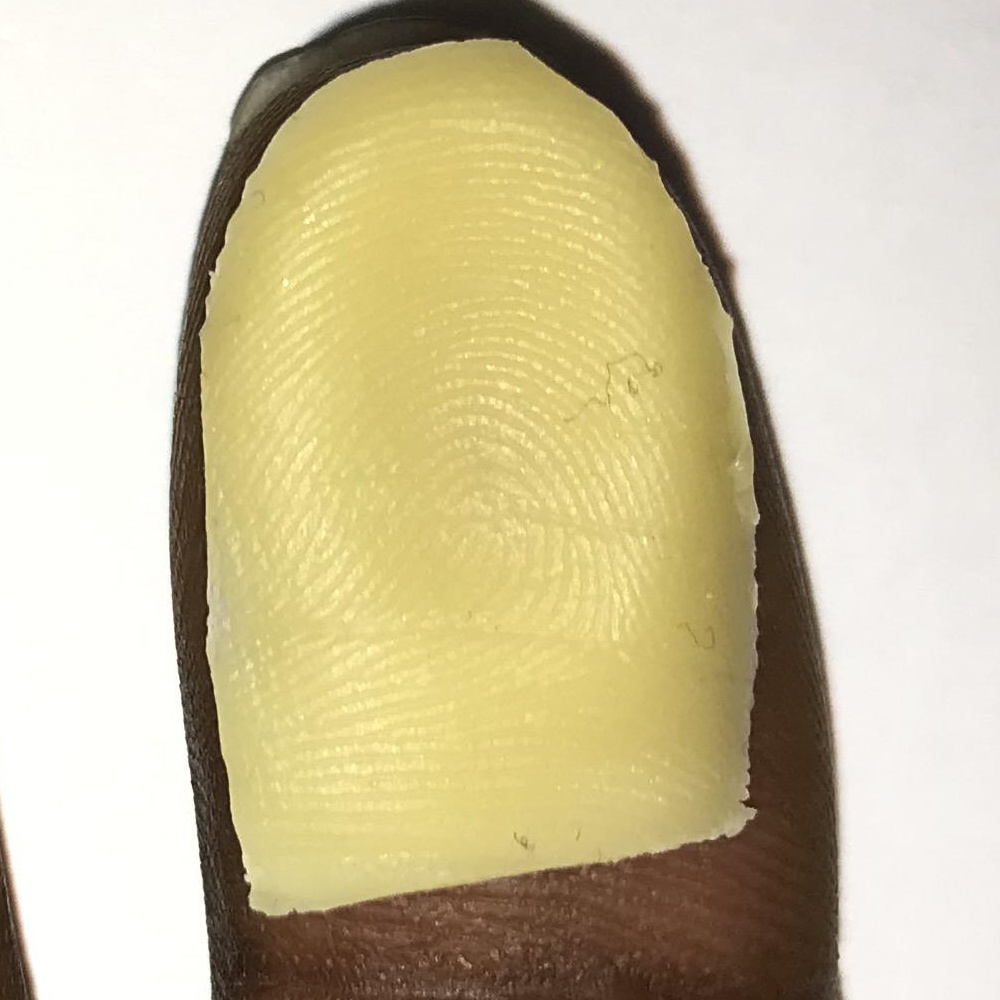}}
\hfill
\subcaptionbox{\centering \small Playdoh PAI}{\includegraphics[height=3cm,width=0.12\textwidth]{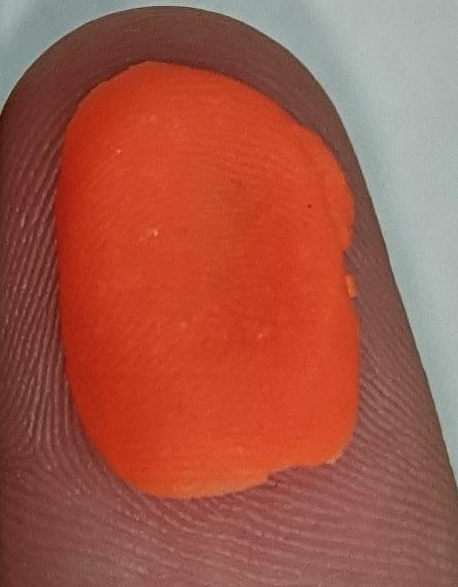}}
\hfill
\subcaptionbox{\centering \small Wood Glue PAI}{\includegraphics[height=3cm,width=0.17\textwidth]{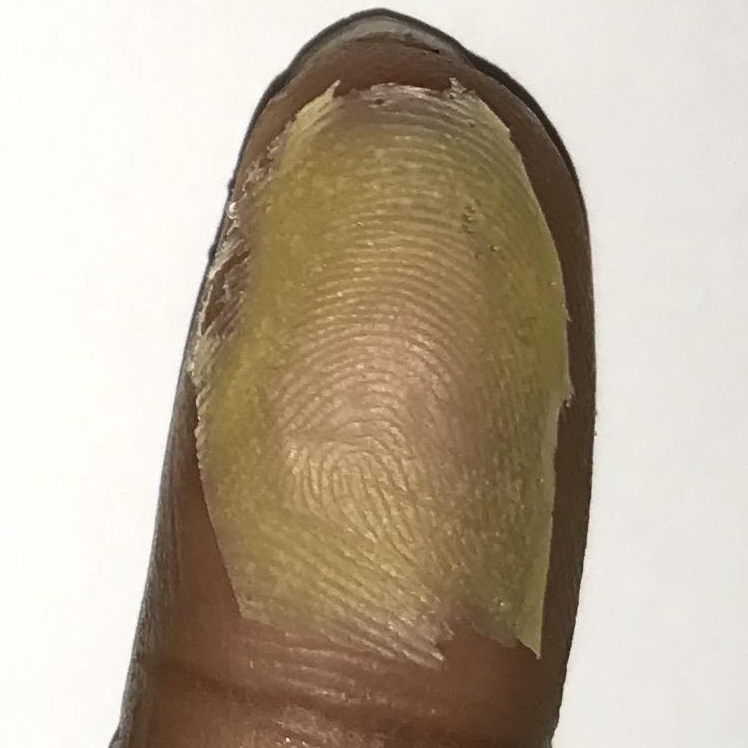}}
\hfill
\subcaptionbox{\centering \small Live}{\includegraphics[height=3cm,width=0.12\textwidth]{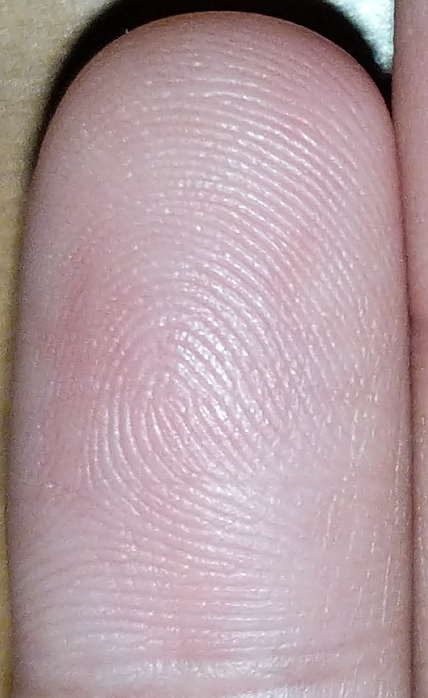}}
\hfill
\subcaptionbox{\centering \small Synthetic Fingertip PAI}{\includegraphics[height=3cm,width=0.11\textwidth]{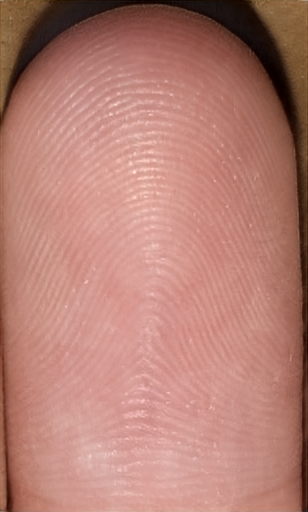}}
\hfill
\caption{\small Example images of all presentation attack types including live samples from the LivDet-2023 Noncontact Fingerprint test dataset}
\label{fig:PAI_Samples}
\end{figure*}

\begin{table*}[!t]
\footnotesize
\centering
\caption{\small LivDet-2023 Noncontact Fingerprint Algorithms Category Dataset Summary}
\label{Dataset}
\begin{tabular}{|c|c|c|c|}
\hline
\textbf{Image type} &\textbf{Training} &\textbf{Testing} &\textbf{Sensors} \\

\hline
Live & 5886 & 2718 & iPhone X*2, iPhone7*2, Samsung Galaxy S6*2 and S7 *2 \\

 &  &  &  Samsung Galaxy S9*2, Google Pixel*2, \\

\hline

Finger Photo PAI & 1104 & 1686 & iPhone X*2, Samsung Galaxy S20, Samsung Galaxy S9, Google Pixel \\
\hline

Ecoflex PAI & 1248 & 1551 & iPhone X*2, iPhone 7*2, Samsung Galaxy S9*2, S20 \\
\hline 

Playdoh PAI & 1623 & 620 & iPhone X*2, iPhone 7*2, Samsung Galaxy S9*2, S20 \\
\hline

Wood Glue PAI & 272 & 2528 & iPhone X*2, iPhone 7*2, Samsung Galaxy S20  \\
\hline

Synthetic PAI & 0 & 1000 & Same as live image  \\

\hline 
Latex PAI & 0 & 2800 & Samsung Galaxy S20 \\

\hline

\multicolumn{4}{c}{Latex PAI and Synthetic PAI have been used as unknown PAIs for algorithm testing}
\end{tabular}
\end{table*}

\begin{table*}[!t]
\footnotesize
\centering
\caption{\small LivDet-2023 Noncontact Fingerprint Systems Category Dataset Summary}
\label{Dataset_Systems}
\begin{tabular}{|c|c|c|c|c|}
\hline

\textbf{Image type} &\textbf{Training} &\textbf{Number of tests on Unified apps} &\textbf{Number of tests on Individual apps} &\textbf{Sensors} \\
\hline

Live & 0 & 270 & 1620 & Samsung Galaxy S9, Samsung Galaxy A71 \\
\hline

Finger Photo PAI & 0 & 72 & 140 & Samsung Galaxy S9, Samsung Galaxy A71 \\
\hline

Ecoflex PAI & 0 & 80 & 400 & Samsung Galaxy S9, Samsung Galaxy A71 \\
 
\hline 
Playdoh PAI & 0 & 68 & 380 & Samsung Galaxy S9, Samsung Galaxy A71 \\
 
\hline
Wood Glue PAI & 0 & 80 & 404 & Samsung Galaxy S9, Samsung Galaxy A71 \\

\hline 
Latex PAI & 0 & 82 & 380 & Samsung Galaxy S9, Samsung Galaxy A71 \\

\hline

\end{tabular}
\end{table*}

\subsection{Evaluation Protocols} The performance of the submissions in both the algorithm and systems category was determined by an output score for each test sample ranging between 0 to 100 with a threshold of 50. A score of 1000 indicates undetected samples. Test samples with scores less than 50 were classified as PAI and scores of 50 and above were classified as live. If the submitted algorithms or systems provided a score of 1000 for the PAIs then it was considered as a correct decision as the algorithm was able to reject PAIs and is not considered as an attack presentation classification error. A score of 1000 for the bonafide sample was considered incorrect and was included as part of the BPCER calculation. All of the evaluations reported in this publication were completed by the competition organizers and were {\it not} self-reported by the competitors.

\section{Competitor's Submission Details} All teams were given the opportunity to submit a description of their submitted algorithm and four such descriptions are provided below. \par
{\textbf{Dermalog:}} Team Dermalog's presentation attack detection algorithm consists of a combination of two CNNs. A patch classifier that scores each patch, and a global CNN that computes weights for each patch score based on the full fingerprint image. The idea is that the patch classifier should focus on the local discriminating pattern and the global CNN should decide the importance of different regions of the image in regard to the problem. The Algorithm starts by resizing the full fingerprint image to 480 x 288 out of which 15 non-overlapping 96x96 patches are cut. The patch classifier scores those patches, which are then weighted with the computed weights of the global CNN. The final output is determined by forwarding the weighted patch scores through a fully connected layer. The patch classifier is a custom residual CNN architecture with $ < $158.000  trainable parameters and the global CNN is a MobileNetV3Small using pre-trained “ImageNet” weights. Team Dermalog trained their model with the training dataset provided for the competition and used a collected in-house dataset of 8952 live and 1276 PAIs (made from silicon, ballistic gelatine, window color, gelatine, playdoh, latex, glue, and paper).

{\textbf{West Virginia University (WVU):}} Team WVU employed an unsupervised learning technique that is specifically trained using only genuine images. When input images are received, a video is generated as input for a 3D Convolutional Neural Network (3DCNN) model to construct a spatial-temporal rPPG block (ST-rPPG block). By performing spatial and temporal sampling from the block, the rPPG signal is extracted. Ultimately, the heart rate from the given input image is determined by computing the power spectral density (PSD) of the signal. To assess the algorithm's performance, datasets containing both authentic and counterfeit finger data are employed. The heart rate obtained from genuine images is expected to fall within the normal range, while that of counterfeit samples will deviate from this range. \par
{\textbf{Fraunhofer:}} The submitted algorithm is based on the attention-based pixel-wise binary supervision network (A-PBS) \cite{fang2021iris}, aiming to capture fine-grained attack patterns. Considering its enhanced generalizability of iris PAD in both NIR and visible spectrum domains \cite{fang2023intra}, the team adapted A-PBS for fingerprint PAD. The goal was to capture subtle patch-level cues by leveraging pixel-wise binary supervision during training and employing an attention mechanism to automatically localize the region that contributes the most to an accurate fingerprint PAD decision. The model utilized two blocks from DenseNet161 and was initialized with the weights trained on the ImageNet dataset. The model was then fine-tuned using Fingerprint PAD data. Specifically, the training data consists of the training set provided by the competition organizers and a publicly available dataset, named COLFISPOOF \cite{kolberg2023colfispoof}. Given the viability of synthetic data in biometrics and its potential as a novel attack type in non-contact fingerprint PAD, we adapted the recently proposed SyPer \cite{kolf2023syper}, which proposed to generate realistic synthetic periocular data, to generate 2500 synthetic fingerprint data for this competition. To ensure diversity in the synthetic fingerprint attack samples, random augmentation techniques such as color transformations, image-space filtering, additive noise, and cutouts were applied during the training of Syper. \par
{\bf Anonymous:} This anonymous team utilized two deep learning models: ResNet 50 and Vision transformer pre-trained on ImageNet as feature extractors. Then the features are passed into the two different SVM models independently. The result obtained from the two SVM models was averaged as the final result to determine the final output.

{\textbf{Tech5 - Systems}} The T5-AirSnap Finger, TECH5’s proprietary contactless finger capture technology, houses the Neural Network (NN) based finger PAD model. Trained on tens of thousands of images, the PAD model can identify a number of 2D and 3D PAIs. It can then generate a liveness score for each detected finger or a unified score for the entire image. The training datasets used are the internally collected dataset and the LivDet-2023 Noncontact Fingerprint training dataset.

{\textbf{Dermalog - Systems}}
The base of the system architecture is a Contactless-Fingerprints app for Android which automatically extracts fingerprints from camera images. The app captures a series of images and segments the fingerprints with a self-trained segmentation model. The segmented fingerprints are evaluated based on their ridge distance. Images are captured until all fingerprints reach a certain quality or a timeout of 10 seconds occurs. Finally, the results are rotated in an upright position and the PAD scores are calculated. For the PAD, Dermalog used the same model that is also used in the algorithm benchmark.

\section{Results and Analysis}
\begin{figure*}[!ht]
\centering

\subcaptionbox{\centering \small  ROC curves for all four algorithms presenting the overall performance on samples representing all six PAIs in the Algorithm competition}{\includegraphics[width=0.48\textwidth]{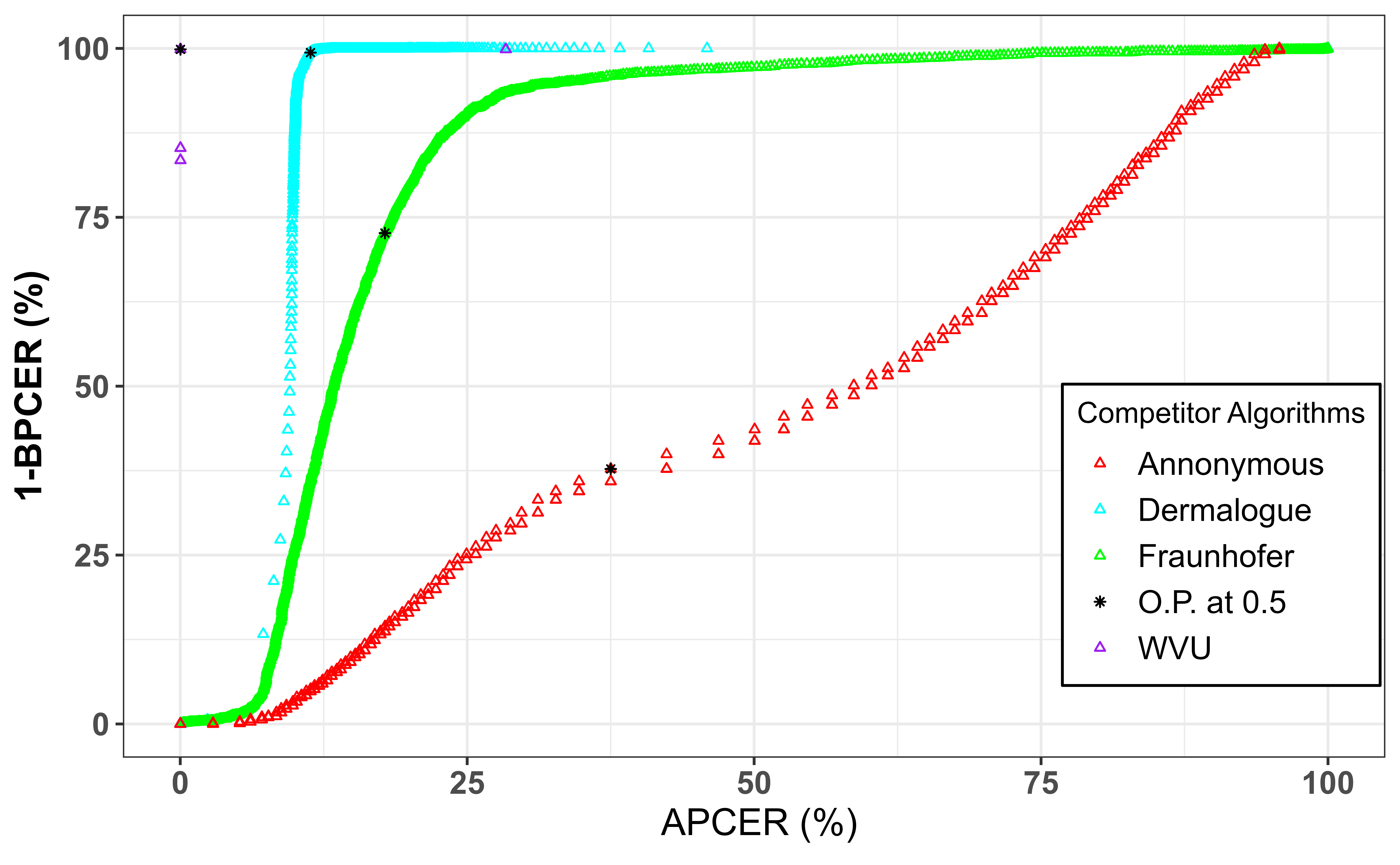}} \label{fig:Algo_ROC}
\hfill
\subcaptionbox{\centering \small ROC curves for all three systems result on two different smartphones presenting the overall performance on samples representing all five PAIs in the System Competition} {\includegraphics[width=0.48\textwidth]{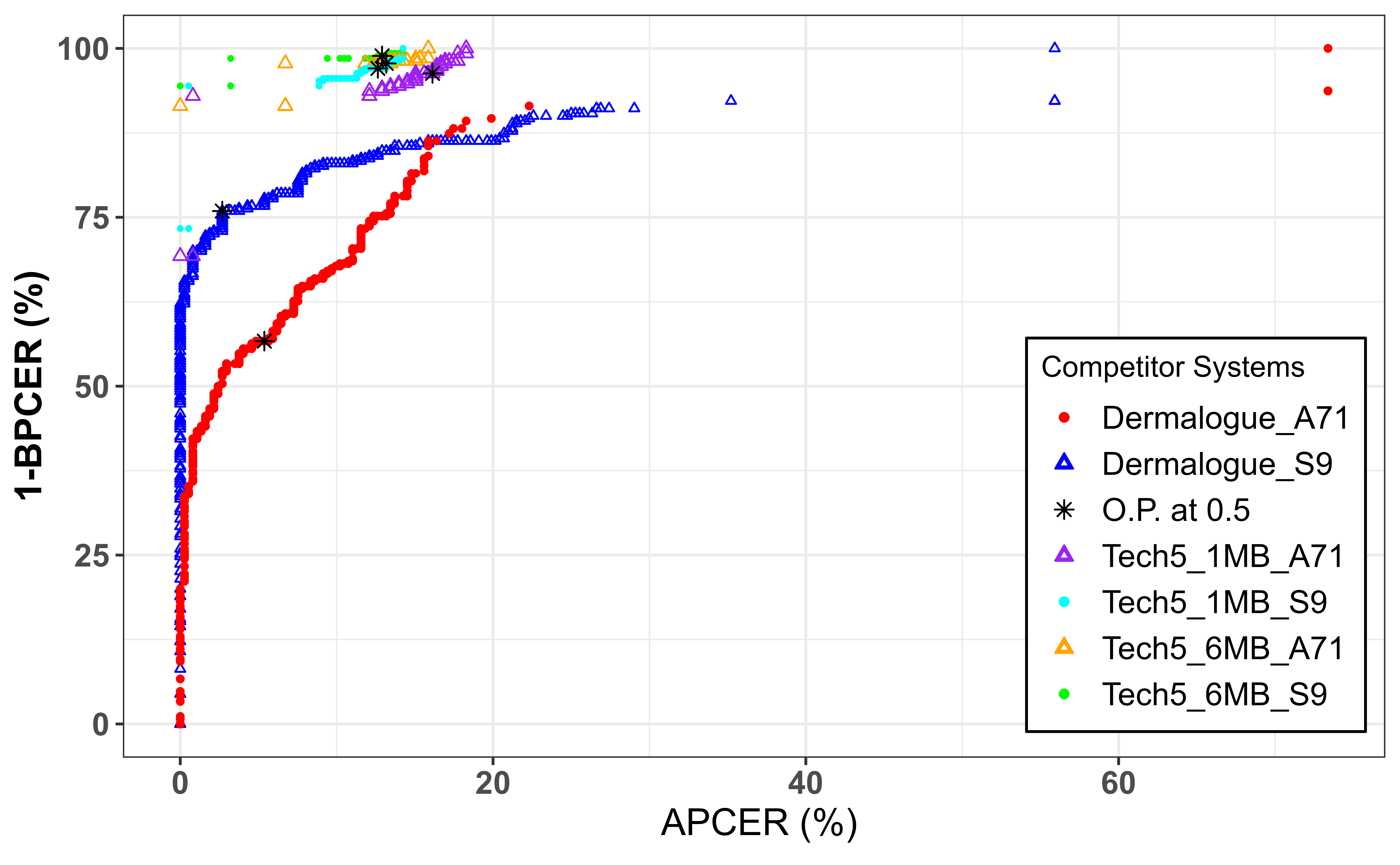}} \label{fig:Sytems_ROC}
\caption{\small ROC curves for LivDet-2023 noncontact fingerprint algorithms and system competition participants. The overall APCER is evaluated based on (APCER$_{\mbox{\footnotesize average}}$). The operating point (``O.P. at 0.5'') used to rank participants of this LivDet-2023 Noncontact Fingerprint Algorithm and System competition is marked by an $\ast$ on each curve.}
\label{fig:ROC_category}

\end{figure*}

This section discusses the results of the LivDet-2023 Noncotact Fingerprint Algorithm and System competition. A summary of the error rates for both the algorithm and system competition is provided in Table \ref{table:Noncontact_Systems_PAD competition results: Algorithms category} and Table \ref{table:Noncontact_Systems_PAD competition results: Systems category}. Additionally, the test results of the individual finger-based noncontact systems are provided in Table \ref{table:Noncontact_Systems_PAD competition results: Individual Systems category}. However, the individual systems were not part of the competition as there was only one competitor in that app category. The performance comparison of the algorithms and the systems category of the competition based on the Receiver Operating Characteristics (ROCs) are shown in Figure~\ref{fig:ROC_category}.

\subsection{Algorithm competition} Team Dermalog is the winner of the algorithms category of the competition with the lowest ACER of 6\%. The second position is occupied by the team WVU with an ACER of 14.28\%. Team WVU achieved the lowest BPCER of 0.15\% among the four competitors in the algorithms category of the competition. The algorithm submitted by the algorithm category winner Dermalog detected all the ecoflex, playdoh, and unknown latex category PAIS successfully. Additionally, team Dermalog achieved APCER 0.10\% against wood glue PAI, which was the lowest for the wood glue PAI modality among all four competitors of the algorithms category. Against the highest difficulty level and unseen (not part of the shared training dataset) synthetic fingertip PAIs, most of the competitors did not perform well, with team WVU achieving the lowest APCER 12.70\% among the competitors with all other competitors scoring APCER 98\% or above against this PAI modality. Team WVU achieved the lowest BPCER of 0.15\% among all competitors in the algorithms category, which led to the second lowest ACER score of 14.28\% to win second place in the competition category. Team Dermalog achieved the lowest APCER$_{\mbox{\footnotesize average}}$ of 11.35\% and the second lowest BPCER of 0.62\% among all four competitors, which helped them win the algorithm competition category. 

To compare the performance of the algorithms for known modalities of PAIs i.e. paper printout, ecoflex, playdoh, and wood glue, the algorithm competition winning team Dermalog performed really well against most PAIs, however, scored APCER of 9.20\% against paper printout PAIs. Similarly, team WVU and Fraunhofer did not perform well against the low difficulty level printout PAIs as well compared to the other PAI modalities. Against the unseen or undisclosed PAI modalities, i.e. latex and synthetic PAIs the winning teams registered a mixed performance. With team Dermalog successfully detecting all latex PAIs. However, three of the competitors did not perform well against synthetic PAIs with all of them registering APCER 98\% or above. Synthetic PAIs can be classified as injection attacks, rather than presentation attacks. This shows the vulnerability of the algorithms against sophisticated injection attacks. 

\begin{table*}[!t]
\footnotesize
\caption{\small LivDet-2023 Noncontact Fingerprint Algorithm competition Summary: Individual Finger Algorithms PAD Results}
\centering
\label{table:Noncontact_Systems_PAD competition results: Algorithms category}
\begin{tabular}{|l|l|l|l|l|l|l|l|l|l|}
\hline

\multirow{4}{*}{\bf Competitor Name} & \multicolumn{6}{c|}{\textbf{Presentation Attack Instruments Level Types}} & \multicolumn{3}{c|}{\bf Overall Performance (\%)}\\
\cline{2-10}

 & {\bf Level A (\%)} & \multicolumn{4}{c|}{\textbf{Level B (\%)}} & {\textbf{Level C (\%)}} & \multirow{3}{*}{\bf APCER$_{\mbox{avg}}$} & \multirow{3}{*}{\bf BPCER} & \multirow{3}{*}{\bf ACER} \\
\cline{2-7}

& {\textbf{Paper printout}} & {\textbf{Ecoflex}} & {\textbf{Playdoh}} & {\textbf{Wood Glue}} &{\textbf{Latex}} & {\textbf{Synthetic fingertip}} & & & \\
\cline{2-10}

\hline\hline

Dermalog & 9.20 & 0 & 0 & 0.10 & 0 & 99.9 & 11.35 & 0.62 & 6  \\
\hline

WVU & 24.60 & 38.10 & 18.60 & 26.90 & 34.43 & 12.70 & 28.40 & 0.15 &  14.28 \\
\hline

Fraunhofer & 33.61 & 0.59 & 0.65 & 1.90 & 7.39 & 98 & 17.83 & 27.31 & 22.57  \\
\hline

Anonymous & 30.80 & 23.92 & 0.48 & 25.79 & 64 & 98.10 & 42.37 & 62.25 & 52.31  \\
\hline

\multicolumn{5}{c}{Latex PAI  and Synthetic PAI have been used as unknown PAIs for testing}
\end{tabular}
\end{table*}

\subsection{System Competition} Although training or validation data was not shared with the system's competition category participants, Tech5 as a member of the CITeR, did have access to the LivDet-2023 Noncontact training dataset. Team Dermalog participated in both competition categories of LivDet-2023 Noncontact Fingerprint Algorithm and System competition and thus had access to the training dataset (a subset of the noncontact PAD dataset) for the algorithm competition category. 

Team Tech5 is the winner of the four-finger-based system competition category with the lowest ACER of 7.36\%. Team Tech5 submitted two unified applications, the Tech5\_6MB\_Unified system achieved the best ACER and was closely followed by the Tech5 second submission, Tech5\_1MB\_Unified with ACER of 8.86\% in the system competition category. Both the systems achieved good performances against moderately difficult levels of PAIs, with the winning system resulting in APCER of 0\% with the A71 smartphone and APCER of 2.50\% with the S9 smartphone against ecoflex, APCER of 0\% with both the A71 and S9 smartphones against playdoh, APCER of 0\% with the A71 smartphone and APCER of 6.25\% with the S9 smartphone against wood glue, and APCER of 0\% with both the A71 and S9 smartphones against unseen latex PAIs. However, both the system's performance was poor against low difficulty level glossy paper printout PAIs. In comparison team Dermalog performed the best against glossy paper printouts with an APCER of 26.40\% with the A71 smartphone and an APCER of 11.12\% with the S9 smartphone. Both the submissions of Tech5 performed well against live samples.Comparatively, team Dermalog achieved a BPCER of 43.34\% with the A71 smartphone and a BPCER of 24.07\% with the S9 smartphone. A better BPCER helped Tech5 win the system competition category compared to team Dermalog. Both the systems performed well against the unknown latex PAIs, which they did not have access to through the competition organizers. \par
For the individual systems, the submitted systems did not perform well compared to the unified systems. Tech5\_1MB\_Individual achieved an APCER of 72.90\% with the A71 smartphone and APCER of 60\% with the S9 smartphone against the lower difficulty PAI level of the glossy paper printout. Similar comparable performances can be noticed against moderately difficult PAIs, including the unseen latex PAI. \par
Furthermore, we have noticed an improvement in performance (for both individual and unified) against the lower difficulty levels of the glossy paper printout PAIs in the older S9 smartphone, compared to the newer A71 smartphone. However, the opposite impact was noticed in the moderately difficult levels of fingertip layover PAIs in the A71 smartphone compared to the older S9 smartphone, which concludes the noncontact fingerprint capture systems installed in newer smartphones are better at layover fingertip PAI detection than older smartphones, probably due to upgrading of camera sensors in the newer smartphone.

\begin{table*}[!t]
\footnotesize
\caption{\small LivDet-2023 Noncontact Fingerprint System Competition Summary: Individual Finger Systems PAD Results}
\centering
\label{table:Noncontact_Systems_PAD competition results: Individual Systems category}
\begin{tabular}{|l|l|l|l|l|l|l|l|l|l|l|l|l|l|l|}
\hline

\multirow{4}{*}{\bf Competitor Name} & \multicolumn{10}{c|}{\textbf{Presentation Attack Instruments Level Types}} & \multicolumn{4}{c|}{\bf Overall Performance (\%)}\\
\cline{2-15}

 & \multicolumn{2}{c|}{\bf Level A (\%)} & \multicolumn{8}{c|}{\textbf{Level B (\%)}} & \multicolumn{2}{c|}{\bf APCER$_{\mbox{avg}}$} & \multicolumn{2}{c|}{\bf BPCER} \\
\cline{2-15}

& \multicolumn{2}{c|}{\textbf{Paper printout}} & \multicolumn{2}{c|}{\textbf{Ecoflex}} & \multicolumn{2}{c|}{\textbf{Playdoh}} & \multicolumn{2}{c|}{\textbf{Wood Glue}} &\multicolumn{2}{c|}{\textbf{Latex}} & & & &  \\
\cline{2-11}

&{\bf A71} & {\bf S9} & {\bf A71} & {\bf S9} &{\bf A71} & {\bf S9} & {\bf A71} & {\bf S9} & {\bf A71} & {\bf S9} &{\bf A71} & {\bf S9} &{\bf A71} & {\bf S9} \\
\hline\hline

Tech5 1MB individual & 72.90 & 60 & 25.50 & 26.93 & 23 & 25.30 & 19.60 & 20 & 22 & 21.32 & 26.63 & 26.40 & 26.22 & 25   \\
\hline

Tech5 6MB individual & 53 & 53 & 23.75 & 12.78 & 23.20 & 22.55 & 28.80 & 43.25 & 20.80 & 19.61 & 26.53 & 27.30 & 24.04 & 21.11  \\
\hline
\multicolumn{5}{c}{Latex PAI has been used as unknown PAIs for testing}
\end{tabular}
\end{table*}

\begin{table*}[!t]
\footnotesize
\caption{\small LivDet-2023 Noncontact Fingerprint System Competition Summary: Four-Finger Systems PAD Results}
\centering
\label{table:Noncontact_Systems_PAD competition results: Systems category}
\begin{tabular}{|l|l|l|l|l|l|l|l|l|l|l|l|l|l|l|l|}
\hline

\multirow{4}{*}{\bf Competitor Name} & \multicolumn{10}{c|}{\textbf{Presentation Attack Instruments Level Types}} & \multicolumn{5}{c|}{\bf Overall Performance (\%)}\\
\cline{2-16}

 & \multicolumn{2}{c|}{\bf Level A (\%)} & \multicolumn{8}{c|}{\textbf{Level B (\%)}} & \multicolumn{2}{c|}{\bf APCER$_{\mbox{avg}}$} & \multicolumn{2}{c|}{\bf BPCER} & \multirow{3}{*}{\bf ACER} \\
\cline{2-15}

& \multicolumn{2}{c|}{\textbf{Paper printout}} & \multicolumn{2}{c|}{\textbf{Ecoflex}} & \multicolumn{2}{c|}{\textbf{Playdoh}} & \multicolumn{2}{c|}{\textbf{Wood Glue}} &\multicolumn{2}{c|}{\textbf{Latex}} & & & & & \\
\cline{2-11}

&{\bf A71} & {\bf S9} & {\bf A71} & {\bf S9} &{\bf A71} & {\bf S9} & {\bf A71} & {\bf S9} & {\bf A71} & {\bf S9} &{\bf A71} & {\bf S9} &{\bf A71} & {\bf S9}& \\
\hline\hline

Tech5\_6MB\_Unified & 68.10 & 57  & 0 & 2.50 & 0 & 0 & 0 & 6.25 &  0 & 0 & 13.17 & 12.90 & 2.23 & 1.12 & 7.36 \\
\hline

Tech5\_1MB\_Unified & 83.34 & 65.30 & 0 & 0 & 0 & 0 & 0 & 0 & 0  & 0 & 16.13 & 12.63 & 3.70 & 2.96 & 8.86  \\
\hline

Dermalog & 26.40 & 11.12  & 0 & 0 & 0 & 2.94 & 0 & 0 & 1.40  & 0 & 5.40 & 2.70 & 43.34 & 24.07 & 18.90  \\
\hline
\multicolumn{5}{c}{Latex PAI has been used as unknown PAIs for testing}
\end{tabular}
\end{table*}

\section{Conclusions}
The Livdet-2023 Noncontact Fingerprint Algorithm and System competition, the first liveness detection competition of the LivDet series, featured multiple new additions to the evaluation of the noncontact fingerprint-based presentation attack detection: (a) employed a novel PAI (synthetic fingertip) in the algorithms category (b) tested the algorithm and systems against unseen PAIs to replicate real-life uncertain scenarios and (c) provided a standard algorithm and system evaluation protocol, along with the comparative analysis of four state-of-the-art algorithms from academia and industry and three state-of-the-art systems submission from the industry. The winning algorithm achieved an ACER of 6\% (APCER averaged over all PAIs = 11.35\% and BPCER = 0.62\%). The winning system of the system competition achieved ACER of 7.36\% (APCER averaged over all PAIs tested and over all A71 and S9 smartphones = 13.04\% and BPCER over all A71 and S9 smartphones = 1.68\%). We have also tested the systems which make the decision on individual finger levels in four-finger settings. Tech5 was the only participant in the individual system category with Tech5\_1MB\_Individual achieved APCER averaged over all PAIs tested and over all A71 and S9 smartphones = 26.48\% and BPCER over all A71 and S9 smartphones = 25.51\% and Tech5\_6MB\_Individual achieved APCER averaged over all PAIs tested and over all A71 and S9 smartphones = 26.48\% and BPCER over all A71 and S9 smartphones = 25.51\%.

From the systems (both unified and individual) results, we have noticed an improvement in performance against the lower difficulty levels of PAIs in the older S9 smartphone compared to the newer A71 smartphone and the opposite in the case of moderately difficult layover PAIs in the newer A71 smartphone compared to the older S9 smartphone. Most of the competitors of the algorithm competition category did not perform well against unseen novel synthetic fingertips or deepfakes with the highest APCER of 99.90\% achieved by the winning algorithm, Dermalog. The competitors with both categories of the competition had direct access to the training dataset to fine-tune their algorithms or systems, still, we see a degradation of performances against the lower difficulty level glossy paper printout PAIs. Our previous work \cite{purnapatra2023presentation} also contributed deep neural network-based architectures like DenseNet-121 and NASNetMobile for PAD using a large subset of the CITeR Noncontact PAD dataset. The DenseNet-121 model achieved APCER$_{\mbox{\footnotesize average}}$ of 0.07\% against known PAIs, a 21.08\% against all PAIs, and BPCER of 0.18\%. While the DenseNet-121 Keras version of the model achieved APCER$_{\mbox{\footnotesize average}}$ of 13.74\% against all PAIs. These results indicate scopes for improvement in PAD performance. The public availability of the PAD dataset would benefit the research community develop more robust PAD solutions in the future.

\bibliographystyle{IEEEtran}
\bibliography{bibliography.bib}

\begin{thebibliography}{10}
\providecommand{\url}[1]{#1}
\csname url@samestyle\endcsname
\providecommand{\newblock}{\relax}
\providecommand{\bibinfo}[2]{#2}
\providecommand{\BIBentrySTDinterwordspacing}{\spaceskip=0pt\relax}
\providecommand{\BIBentryALTinterwordstretchfactor}{4}
\providecommand{\BIBentryALTinterwordspacing}{\spaceskip=\fontdimen2\font plus
\BIBentryALTinterwordstretchfactor\fontdimen3\font minus
  \fontdimen4\font\relax}
\providecommand{\BIBforeignlanguage}[2]{{%
\expandafter\ifx\csname l@#1\endcsname\relax
\typeout{** WARNING: IEEEtran.bst: No hyphenation pattern has been}%
\typeout{** loaded for the language `#1'. Using the pattern for}%
\typeout{** the default language instead.}%
\else
\language=\csname l@#1\endcsname
\fi
#2}}
\providecommand{\BIBdecl}{\relax}
\BIBdecl

\bibitem{lin2018matching}
C.~Lin and A.~Kumar, ``Matching contactless and contact-based conventional
  fingerprint images for biometrics identification,'' \emph{IEEE Transactions
  on Image Processing}, pp. 2008--2021, 2018.

\bibitem{labati2015toward}
R.~D. Labati, A.~Genovese, V.~Piuri, and F.~Scotti, ``Toward unconstrained
  fingerprint recognition: A fully touchless 3-d system based on two views on
  the move,'' \emph{IEEE transactions on systems, Man, and cybernetics:
  systems}, 2015.

\bibitem{fujio2018face}
M.~Fujio, Y.~Kaga, T.~Murakami, T.~Ohki, and K.~Takahashi, ``Face/fingerphoto
  spoof detection under noisy conditions by using deep convolutional neural
  network.'' in \emph{BIOSIGNALS}, 2018, pp. 54--62.

\bibitem{marasco2021fingerphoto}
E.~Marasco and A.~Vurity, ``Fingerphoto presentation attack detection:
  Generalization in smartphones,'' in \emph{2021 IEEE International Conference
  on Big Data (Big Data)}.\hskip 1em plus 0.5em minus 0.4em\relax IEEE, 2021.

\bibitem{schuckers2023fido}
S.~Schuckers, G.~Cannon, and N.~Tekampe, ``{FIDO Biometrics Requirements},''
  \url{https://fidoalliance.org/specs/biometric/requirements/Biometrics-Requirements-v3.0-fd-20230111.html},
  2021, accessed: 2023-01-19.

\bibitem{ISO_IEC_301073:2017}
{ISO/IEC 30107-3}, ``{Information technology -- Biometric presentation attack
  detection -- Part 3: Testing and reporting},'' 2016.

\bibitem{lee2006preprocessing}
C.~Lee, S.~Lee, J.~Kim, and S.-J. Kim, ``Preprocessing of a fingerprint image
  captured with a mobile camera,'' in \emph{International conference on
  biometrics}.\hskip 1em plus 0.5em minus 0.4em\relax Springer, 2006, pp.
  348--355.

\bibitem{lee2008recognizable}
D.~Lee, K.~Choi, H.~Choi, and J.~Kim, ``Recognizable-image selection for
  fingerprint recognition with a mobile-device camera,'' \emph{IEEE
  Transactions on Systems, Man, and Cybernetics, Part B (Cybernetics)},
  vol.~38, no.~1, pp. 233--243, 2008.

\bibitem{li2012testing}
G.~Li, B.~Yang, R.~Raghavendra, and C.~Busch, ``Testing mobile phone camera
  based fingerprint recognition under real-life scenarios,'' \emph{NISK},
  vol.~1, p.~2, 2012.

\bibitem{sankaran2015smartphone}
A.~Sankaran, A.~Malhotra, A.~Mittal, M.~Vatsa, and R.~Singh, ``On smartphone
  camera based fingerphoto authentication,'' in \emph{2015 IEEE 7th
  International Conference on Biometrics Theory, Applications and Systems
  (BTAS)}.\hskip 1em plus 0.5em minus 0.4em\relax IEEE, 2015.

\bibitem{jawade2021multi}
B.~Jawade, A.~Agarwal, S.~Setlur, and N.~Ratha, ``Multi loss fusion for
  matching smartphone captured contactless finger images,'' in \emph{2021 IEEE
  International Workshop on Information Forensics and Security (WIFS)}.\hskip
  1em plus 0.5em minus 0.4em\relax IEEE, 2021, pp. 1--6.

\bibitem{jawade2022ridgebase}
B.~Jawade, D.~D. Mohan, S.~Setlur, N.~Ratha, and V.~Govindaraju, ``Ridgebase: A
  cross-sensor multi-finger contactless fingerprint dataset,'' in \emph{2022
  IEEE International Joint Conference on Biometrics (IJCB)}.\hskip 1em plus
  0.5em minus 0.4em\relax IEEE, 2022, pp. 1--9.

\bibitem{stein2013video}
C.~Stein, V.~Bouatou, and C.~Busch, ``Video-based fingerphoto recognition with
  anti-spoofing techniques with smartphone cameras,'' in \emph{2013
  International Conference of the BIOSIG Special Interest Group
  (BIOSIG)}.\hskip 1em plus 0.5em minus 0.4em\relax IEEE, 2013.

\bibitem{taneja2016fingerphoto}
A.~Taneja, A.~Tayal, A.~Malhorta, A.~Sankaran, M.~Vatsa, and R.~Singh,
  ``Fingerphoto spoofing in mobile devices: a preliminary study,'' in
  \emph{2016 IEEE 8th International Conference on Biometrics Theory,
  Applications and Systems (BTAS)}.\hskip 1em plus 0.5em minus 0.4em\relax
  IEEE, 2016.

\bibitem{wasnik2018presentation}
P.~Wasnik, R.~Ramachandra, K.~Raja, and C.~Busch, ``Presentation attack
  detection for smartphone based fingerphoto recognition using second order
  local structures,'' in \emph{2018 14th International Conference on
  Signal-Image Technology \& Internet-Based Systems (SITIS)}.\hskip 1em plus
  0.5em minus 0.4em\relax IEEE, 2018.

\bibitem{kolberg2023colfispoof}
J.~Kolberg, J.~Priesnitz, C.~Rathgeb, and C.~Busch, ``Colfispoof: A new
  database for contactless fingerprint presentation attack detection
  research,'' in \emph{Proceedings of the IEEE/CVF Winter Conference on
  Applications of Computer Vision}, 2023.

\bibitem{priesnitz2022syncolfinger}
J.~Priesnitz, C.~Rathgeb, N.~Buchmann, and C.~Busch, ``Syncolfinger: Synthetic
  contactless fingerprint generator,'' \emph{Pattern Recognition Letters}, vol.
  157, pp. 127--134, 2022.

\bibitem{purnapatra2023presentation}
S.~Purnapatra, C.~Miller-Lynch, S.~Miner, Y.~Liu, K.~Bahmani, S.~Dey, and
  S.~Schuckers, ``Presentation attack detection with advanced cnn models for
  noncontact-based fingerprint systems,'' \emph{arXiv preprint
  arXiv:2303.05459}, 2023.

\bibitem{LivDet}
{LivDet Organizing Team}, ``Livdet website,'' available at: http://livdet.org/.

\bibitem{fang2021iris}
M.~Fang, N.~Damer, F.~Boutros, F.~Kirchbuchner, and A.~Kuijper, ``Iris
  presentation attack detection by attention-based and deep pixel-wise binary
  supervision network,'' in \emph{2021 IEEE International Joint Conference on
  Biometrics (IJCB)}.\hskip 1em plus 0.5em minus 0.4em\relax IEEE, 2021, pp.
  1--8.

\bibitem{fang2023intra}
M.~Fang, F.~Boutros, and N.~Damer, ``Intra and cross-spectrum iris presentation
  attack detection in the nir and visible domains,'' in \emph{Handbook of
  Biometric Anti-Spoofing: Presentation Attack Detection and Vulnerability
  Assessment}.\hskip 1em plus 0.5em minus 0.4em\relax Springer, 2023, pp.
  171--199.

\bibitem{kolf2023syper}
J.~N. Kolf, J.~Elliesen, F.~Boutros, H.~Proen{\c{c}}a, and N.~Damer, ``Syper:
  Synthetic periocular data for quantized light-weight recognition in the nir
  and visible domains,'' \emph{Image and Vision Computing}, vol. 135, p.
  104692, 2023.

\end{thebibliography}

\end{document}